\documentclass[sigconf, nonacm]{acmart}
\AtBeginDocument{%
  \providecommand\BibTeX{{%
    \normalfont B\kern-0.5em{\scshape i\kern-0.25em b}\kern-0.8em\TeX}}}

\definecolor{mygray}{gray}{.9}
\definecolor{mygray1}{gray}{.7}
\usepackage{colortbl}
\usepackage{color}
\usepackage{multirow}
\usepackage{graphicx}
\usepackage{scalerel}
\usepackage{bm}

\definecolor{yellow}{rgb}{0.75, 0.76, 0.02}
\definecolor{orange}{rgb}{0.89, 0.46, 0.29}

\makeatletter
\newcommand{\thickhline}{%
	\noalign {\ifnum 0=`}\fi \hrule height 1pt
	\futurelet \reserved@a \@xhline
}
\makeatother
\begin{document}

\title{\texttt{AudioScenic}: Audio-Driven Video Scene Editing}


\author{Kaixin Shen, Ruijie Quan, Linchao Zhu, Jun Xiao, Yi Yang\\
Zhejiang University}

% \author{Name}
% \affiliation{%
%   \institution{Institution}
%   % \streetaddress{1 Th{\o}rv{\"a}ld Circle}
%   \city{City}
%   \country{Country}}
% \email{xx@xx.xx}

% \author{Name}
% \affiliation{%
%   \institution{Institution}
%   % \streetaddress{1 Th{\o}rv{\"a}ld Circle}
%   \city{City}
%   \country{Country}}
% \email{xx@xx.xx}

% \renewcommand{\shortauthors}{author name and author name, et al.}

\begin{abstract}
  Audio-driven visual scene editing endeavors to manipulate the visual background while leaving the foreground content unchanged, according to the given audio signals. Unlike current efforts focusing primarily on image editing, audio-driven video scene editing has not been extensively addressed.
 In this paper, we introduce \texttt{AudioScenic}, an audio-driven framework designed for video scene editing. 
\texttt{AudioScenic} integrates audio semantics into the visual scene through a temporal-aware audio semantic injection process. As our focus is on background editing, we further introduce a SceneMasker module, which maintains the integrity of the foreground content during the editing process.
\texttt{AudioScenic} exploits the inherent properties of audio, namely, audio magnitude and frequency, to guide the editing process, aiming to control the temporal dynamics and enhance the temporal consistency. 
First, we present an audio Magnitude Modulator module that adjusts the temporal dynamics of the scene in response to changes in audio magnitude, enhancing the visual dynamics. Second, the audio Frequency Fuser module is designed to ensure temporal consistency by aligning the frequency of the audio with the dynamics of the video scenes, thus improving the overall temporal coherence of the edited videos.
These integrated features enable \texttt{AudioScenic} to not only enhance visual diversity but also maintain temporal consistency throughout the video. 
We present a new metric named temporal score for more comprehensive validation of temporal consistency. We demonstrate substantial advancements of \texttt{AudioScenic} over competing methods on DAVIS~\cite{davis2017} and Audioset~\cite{audioset} datasets.
\end{abstract}

% <concept_id>10010147.10010178.10010224</concept_id>
% <concept_desc>Computing methodologies~Computer vision</concept_desc>
% <concept_significance>300</concept_significance>

% \ccsdesc[300]{Computing methodologies~Computer vision}




\maketitle

\section{Introduction}
\begin{figure*}[t]
  \centering
  \includegraphics[width=1.0\textwidth]{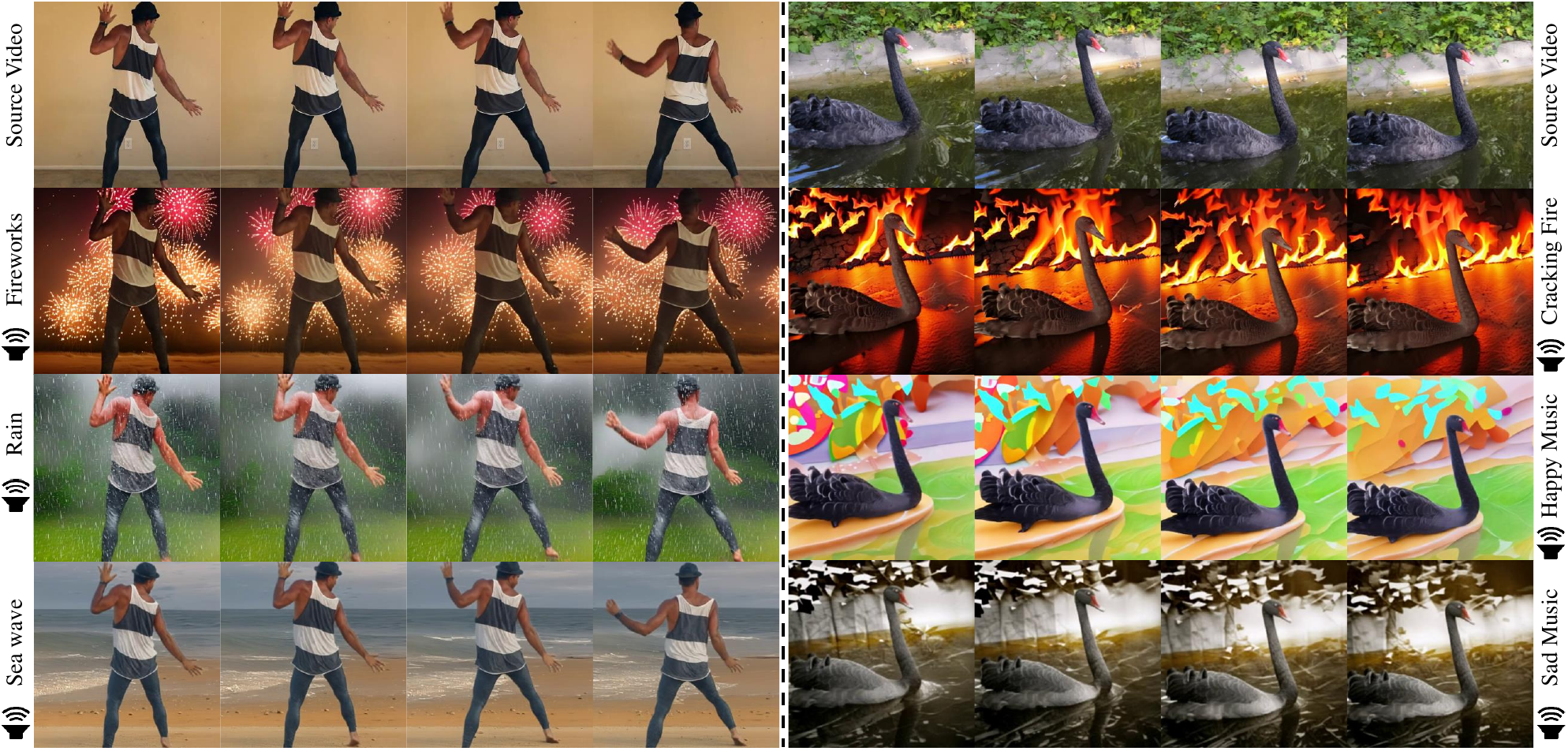}
 \captionof{figure}{
    \textbf{Editing results of \texttt{AudioScenic}.} Given a source video (top row), our approach can perform scene editing using various audio clips while preserving the foreground content. Moreover, we can conduct scene-style transitions within emotional audio clips like happy music or sad music.
    }
  \label{Fig.figure1}
  \vspace{-10pt}
\end{figure*}
\label{sec:intro}

Visual scene editing aims to manipulate the visual background content while keeping the foreground objects unaltered, which is crucial in real-world applications, such as TikTok videos.
%serving as the backbone of storytelling across various platforms, from movies and television to online content. 
Recent audio-driven works primarily focus on using diffusion models~\cite{ddpm, scorebaedddpm, ddim, latentdiffusion, hierarchical, UGC, SDXL, DiffusionGPT} to guide image editing~\cite{lee2020crossing,soundguided}, utilizing a source image as input and conducting editing on a static image given the audio conditions. 

In this paper, we introduce a new setting, audio-driven video scene editing, aiming to dynamically adapt the visuals to the changing tones and semantics of the audio.
%Previous methods~\cite{lee2020crossing,soundguided} mainly centered on audio-conditioned image scene editing, accepting a source image as input and facilitating scene editing within the audio semantic condition. 
When utilizing audio to edit video scenes, three principal challenges need to be addressed: \textbf{i)} Foreground preservation: while editing the scene areas of videos, the foreground content should remain unaltered. \textbf{ii)} Audio-aligned temporal dynamics: after accurately editing the scene content, it is essential to present temporal visual variations aligned with the audio condition (e.g., in ``Rain'' scenes, the intensity of the rain changes over time and varies with the fluctuations in rainy sound). Neglecting temporal dynamics can result in static scenes in the edited frames. \textbf{iii)} Temporal consistency: maintaining temporal consistency is crucial in video editing to ensure consistency across frames and prevent visual flickering. 

To address these challenges, we introduce \texttt{AudioScenic}, a new framework that leverages audio as a condition to guide video scene editing. In particular, \texttt{AudioScenic} includes the temporal-aware audio semantics guidance process and three specialized modules: SceneMasker, Magnitude Modulator, and Frequency Fuser. Inside the temporal-aware audio semantic injection process, we extract semantic embeddings from audio clips, element-wise adding them with timestep embedding derived from the sampling timestep. Then we utilize the timestep embedding fused with audio semantics to direct a latent denoising process within an adapted 3D U-Net diffusion model during both training and inference stage. 1) \textbf{To preserve foreground content}, we propose to utilize SceneMasker, a mask blending module, to restrict the audio embedding's influence exclusively to the scene areas of videos. In addition, relying solely on audio semantics is insufficient to create temporally dynamic and coherent video scenes. We observe that magnitude and frequency are pivotal audio properties in this regard. 2) \textbf{To create temporal dynamics}, we employ the audio magnitude as a controller. The audio magnitude plays a crucial role in modulating the fine-grained temporal variations during the scene editing~\cite{aadiff}. For instance, a stronger audio magnitude can lead to more dramatic changes in the video scene, enabling a broader spectrum of visual dynamics. Consequently, we design a module named Magnitude Modulator to modulate the influence of audio semantic embeddings on the video based on the audio magnitude features. 3) Further, we employ audio frequency information \textbf{to maintain temporal consistency}. Audio and video are naturally aligned in the frequency domain along the time axis, thus frequency information contains essential domain conditions of the audio data, which provide valuable temporal references~\cite{lin2024echotrack}. We introduce Frequency Fuser, which merges audio and video frequency information. It transforms spatial video features into frequency space, enabling their integration with audio frequency features through a weighted-multiplication mechanism. 

On the whole, \texttt{AudioScenic} demonstrates several compelling features: \textbf{First}, it combines audio guidance and SceneMasker module for video scene editing, striking a delicate balance by editing the scene while keeping the foreground content unaltered. Audio clips with the same semantic label can be employed to generate varying visual scenes. We can also utilize emotional resonance like music mood to conduct scene style editing, as shown in Fig \ref{Fig.figure1}. \textbf{Second}, \texttt{AudioScenic} leverages the Magnitude Modulator module to control the scene content based on the audio magnitude. Such magnitude control enhances the temporal dynamics of synthesized scenes. \textbf{Third}, the Frequency Fuser module inside \texttt{AudioScenic} enables the model to focus on frequency elements that demonstrate a true cross-modal correlation between audio conditions and video scenes over time, making the results maintain temporal consistency.

For more comprehensive validation of temporal consistency, we present a new metric named temporal score. The temporal score measures the temporal consistency of results on the premise of semantically accurate scene editing. \textbf{By fully exploiting the attributes of audio such as semantics, magnitude, and frequency}, the result of \texttt{AudioScenic} is the generation of scenes that are not only coherent but also exhibit high diversity.

In a nutshell, our contributions are three-fold:
\begin{itemize}
  \item We introduce a new task termed audio-driven video scene editing and develop a comprehensive protocol with a new evaluation metric, the temporal score, to enhance the assessment of methodological performance. 
  \item We propose an audio-driven framework named \texttt{AudioScenic} to edit video scenes. \texttt{AudioScenic} utilizes audio magnitude and frequency information to enhance temporal dynamics and ensure temporal consistency in video scene editing. 
  \item We demonstrate promising applications in video editing, showing that our method produces audio-synchronized videos and outperforms previous text-driven and audio-driven methods in video scene editing, as demonstrated by video samples using the DAVIS~\cite{davis2017} and Audioset~\cite{audioset} datasets.
\end{itemize}

\section{Related Work}
\label{sec: related work}

\textbf{Text-Driven Video Editing.} Compared to image editing~\cite{prompttoprompt, blendeddiffusion, anydoor, Coarse-to-Fine, RePaint, GLIDE, ImagenEditor, SmartBrush, DreamInpainter, PaintbyExample, Inst-Inpaint}, video editing~\cite{Style-A-Video, Talking-head, FLATTEN, Off-The-Shelf, CustomizingMotion, MotionDirector, VMC, VideoDreamer} is more challenging due to its extra temporal dimension. Tune-a-Video~\cite{Tune} first inflates a text-to-image 2D diffusion model for video editing, it finetunes the model on a single video and generates new videos. Fatezero~\cite{fatezero} takes inspiration from Prompt-to-Prompt~\cite{prompttoprompt} and edits the videos by changing the text-image cross-attention map, showing promising results in preserving foreground shape and motion. Dreamix~\cite{dreamix} uses a text-to-video backbone for motion editing and preserves temporal consistency. Both Text2Live~\cite{text2live} and StableVideo~\cite{stablevideo} utilize Layered Neural Atlases. Text2Live divides the video into several layers and edits each layer separately through a text description. StableVideo designs an inter-frame propagation mechanism and aggregation network to generate the edited atlases from the keyframes, thus achieving temporal and spatial consistency. Ground-a-Video~\cite{grouond} extracts the layout of objects in the video and uses this location condition for editing. TokenFlow\cite{xing2024tokenflow} centers on enhancing video latent feature smoothness to reduce the video visual flickering. Rerender-A-Video~\cite{yang2023rerender} proposes a hierarchical cross-frame constraint to preserve temporal consistency. UniEdit~\cite{bai2024uniedit} aims to achieve zero-shot motion and texture editing by injecting conditions into self-attention and cross-attention layers. Solely using text for editing content is unsuitable for indescribable objects and scenes, prompting recent work to leverage auxiliary visual conditions. We experimentally observed that text-based methods may face challenges in editing video scenes. 

We extend the condition modality to audio and improve the video scene synthesis quality using audio magnitude and frequency information since both audio and video naturally have temporal information, which is lacking in other modalities like text.

\noindent\textbf{Audio-Guided Visual Synthesis.} Prior methods mainly leverage audio to conduct image generation or video generation. Several methods~\cite{soundgeneration,audiotoken} focus on image synthesis using generative models. ~\cite{soundguided} employs audio to edit images. Recently some works have explored audio-driven video generation. AAdiff~\cite{aadiff} combines Prompt-to-Prompt~\cite{prompttoprompt} with audio control, generating videos by multiplying the audio magnitude and the corresponding text token embedding. TPoS~\cite{tpos} integrates audio with temporal semantics and magnitude, generating a specified image and then conducting image-to-video generation. ~\cite{xing2024seeing} realizes audio-to-video, video-to-audio, and audio-video joint generation using audio and text as conditions. However, they rely on text-relevant audio and utilize audio as a textual auxiliary condition to generate videos through 2D text-to-image diffusion models.
Unlike most existing audio-driven works conducting visual generation, we aim to employ audio to guide video scene editing.
Sound-G~\cite{soundguided} utilizes StyleGAN~\cite{karras2020stylegan} to edit images using audio semantics. Due to the absence of temporal information, it is incapable of editing videos. The utilization of StyleGAN also results in its poor generalizability. Within audio semantic information, 
We additionally incorporate the audio magnitude and frequency feature to present temporal consistent videos. We show our ability to generate diverse and dynamic scenes.

% Soundini~\cite{soundini} achieves local region editing within audio and user-provided visual masks. It is only suitable for editing natural videos without foreground objects. Also, it does not utilize the audio frequency feature, which includes noteworthy temporal information. We conduct global scene editing and focus on editing broader videos containing both complex foreground and scene content. By fully using audio semantic and magnitude information, We additionally incorporate the audio frequency feature to present more consistent videos. We show our ability to synthesize temporal coherent and dynamic scenes while keeping foreground content unchanged.

\section{Preliminaries}
\label{sec:formatting}

\begin{figure*}[t]
  \centering
  \includegraphics[width=1.0\textwidth]{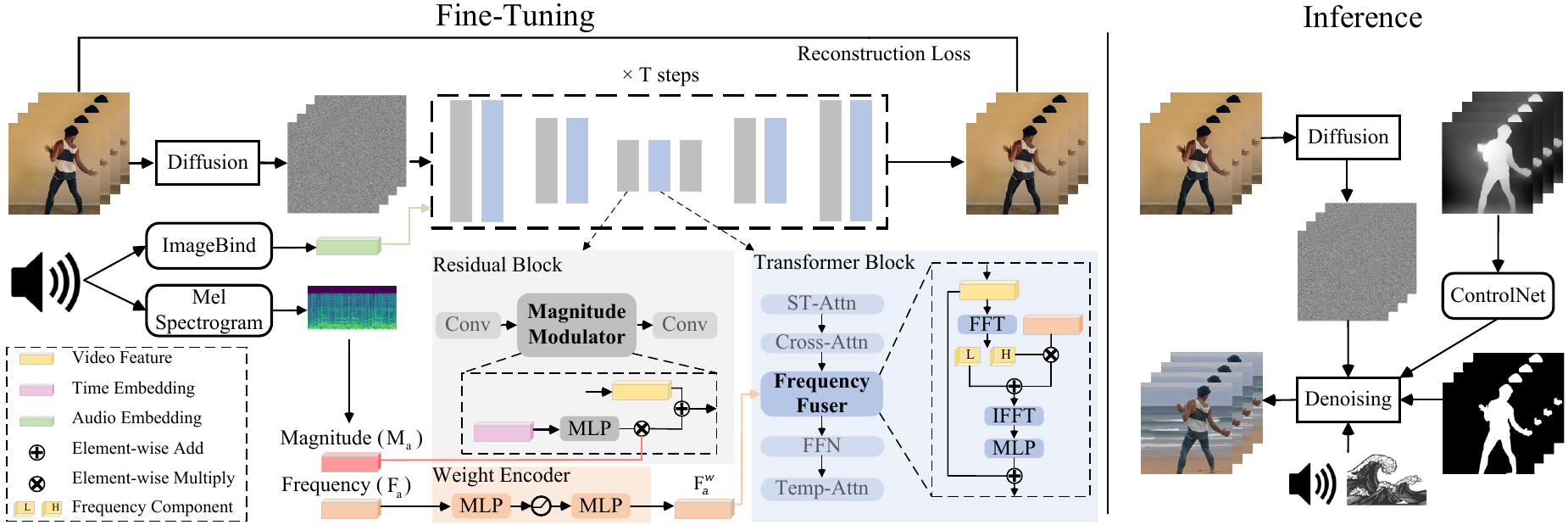}
  \caption{Framework of \texttt{AudioScenic} (\S\ref{sec: framework}). In the fine-tuning stage, given a source video within an audio clip, we invert the video to noisy latent using DDIM inversion. We obtain semantic embedding, magnitude feature, and frequency feature from the audio clip. We fuse audio semantic embedding with timestep embedding derived from timestep $t$ for guiding the latent denoising process. The audio magnitude feature $\bm{M}_a$ and frequency feature $\bm{F}_a$ are used for the Magnitude Modulator and the Frequency Fuser, respectively. We compute the reconstruction loss for fine-tuning. In the Inference stage, we input a new audio clip for guidance. The depth and binary masks are additionally used to preserve the foreground content.}
  \label{Fig.main}
  \vspace{-10pt}
\end{figure*}

\textbf{Stable Diffusion.} In this paper, we use Stable Diffusion, a widely used text-to-image model, to achieve video editing. Stable Diffusion is based on the Latent Diffusion Model (LDM), which conducts the denoising process in the latent space of an autoencoder, namely $\mathcal{E}(\cdot)$ and $\mathcal{D}(\cdot)$, implemented via VAE pre-trained on large image datasets. This design shows an advantage in reducing computational costs while keeping the visual quality. 

During the training stage, an input image $\bm{x}_0 \in \mathbb{R}^{H \times W \times 3}$ is initially compressed to the latent space by the frozen encoder, yielding $\bm{z}_0\!=\!\mathcal{E}(\bm{x}_0)$, then diffusion forward process gradually adds Gaussian noise to $\bm{z}_{0}$ to obtain $z_{t}$ through Markov transition with the transition probability:
\begin{equation}\small
    q(\bm{z}_t | \bm{z}_{t-1}) = \mathcal{N}(\bm{z}_t; \sqrt{1-\beta_t}\bm{z}_{t-1}, \beta_t\mathit{I}),
\end{equation}
for $t = 1,\ldots, T$, with $T$ denoting the number of diffusion
timestep. The sequence of hyper-parameters $\beta_t$ determines the noise strength at each step. Then, the backward denoising process is given by the transition probability:
\begin{equation}\small
\begin{aligned}
p_{\theta}(\bm{z}_{t-1}|\bm{z}_{t}) = \mathcal{N}(\bm{z}_{t-1}; \mu_{\theta}(\bm{z}_{t},t), \sigma^{2}_{t}\textit{I})\:,
\end{aligned}
\end{equation}
for $t = T,\ldots, 1$. Here the mean $\mu_{\theta}(\bm{z}_{t},t)$ can be represented using the noise predictor $\epsilon_{\theta}$ which is learned by the minimization of the MSE loss of network parameter $\theta$:
\begin{equation}\small
    \mathcal{L} = \mathbb{E}_{\mathcal{E}(\bm{x}_0), y, \epsilon\sim\mathcal{N}(0, \mathit{I}), t}\left\lbrack 
\lVert \epsilon - \epsilon_\theta(\bm{z}_t, t, \tau_\theta(y)) \rVert_2^2 \right\rbrack,
\end{equation}
where $y$ is the corresponding textual description, $\tau_\theta(\cdot)$ is a text encoder mapping the text $y$ to an embedding.

% Specifically, a prevalent approach in diffusion-based image editing is to use the deterministic DDIM scheme to accelerate the sampling process. Within this method, the noisy latent $z_{T}$ can be transformed into a fully denoised latent $\bm{z}_{0}$: %with a reduced number of timesteps:
% \begin{equation}\small
% \begin{aligned}
% \bm{z}_{t-1} = \sqrt{\cfrac{\alpha_{t-1}}{\alpha_{t}}} \bm{z}_{t} 
% + \left( \sqrt{\cfrac{1-\alpha_{t-1}}{\alpha_{t-1}}}
% - \sqrt{\cfrac{1-\alpha_{t}}{\alpha_{t}}} \right) \epsilon_{\theta},
% \end{aligned}
% \end{equation}
% for $t = T,\ldots, 1$. $\alpha_{t}$ is a reparameterized noise scheduler.

\noindent\textbf{Audio Preprocessing.} We take audio clips $A$ as conditional. We introduce audio preprocessing for extracting audio semantic embedding $\bm{E}_a$, magnitude features $\bm{M}_a$, and frequency features $\bm{F}_a$. 

To obtain the audio semantic embedding $\bm{E}_a$, we employ the ImageBind~\cite{imagebind}, a multi-modality model to extract the embedding from audio, thereby eliminating the need for additional training. 

For acquiring the magnitude feature $\bm{M}_a$, we first employ a sliding window to divide the input audio clip into $N$ chunks. Note that $N$ corresponds to the number of frames along the time axis. Then we calculate the average magnitude within each chunk. Subsequently, we normalize and smooth the averaged magnitude of each chunk using the Softmax function:
\begin{equation}\small
\begin{aligned}
\text{Softmax}(\bm{M}_i) = \frac{\text{exp}(\bm{M}_i/\tau)}{\sum_{n \in N}\text{exp}(\bm{M}_n/\tau)},
\end{aligned}
\end{equation}
where $\bm{M}_i$ denotes the $i$th chunk. $\tau$ is the temperature parameter, controlling the smoothness of magnitude value of the audio chunks. The smoothness prevents the magnitude changes of the audio temporally drastic. Now we obtain audio magnitude $\bm{M}_a \!=\! \{\bm{M}_i\}_1^N$.

To extract the audio frequency feature $\bm{F}_a$, we utilize a Mel Spectrogram (Mel). This method applies a frequency-domain filter bank to time-windowed audio clips and outputs the audio frequency feature, expressed as $\bm{F}_{a}\!=\! \text{Mel}(A)$.

\section{Method}
\textbf{Problem Formulation.} Given the source video $V \in \mathbb{R}^{N \times C \times W \times H}$ and source audio clip $A \in \mathbb{R}^{L}$, our goal is to predict the scene-edited video $V^{*} \in \mathbb{R}^{N \times C \times W \times H}$ with a new audio clip $A^{*} \in \mathbb{R}^{L}$, where $N$ is the number of frames, $C, W, H$ denote the channel, width, and height, respectively, $L$ is the audio clip duration.

\noindent\textbf{Overview.} Our \texttt{AudioScenic} focuses on editing video scenes (backgrounds) with the assistance of audio properties such as semantics, magnitude, and frequency. We utilize a modified version of Stable Diffusion as our foundational framework for scene editing (\S\ref{sec: framework}). We propose a temporal-aware audio semantic injection process to integrate audio semantics into the model for editing guidance (\S\ref{sec: audiocondition}). To edit the scenes while keeping the foreground content unaltered, we employ SceneMasker to restrict the impact of audio conditions exclusively to the scene areas (\S\ref{scenemasker}). Moreover, we introduce two audio-specific modules: Magnitude Modulator (\S\ref{magnitude modulator}) and Frequency Fuser (\S\ref{frequency fuser}), aimed at enhancing temporal dynamics and ensuring temporal consistency, respectively.

\subsection{\texttt{AudioScenic} Framework}
\label{sec: framework}
% We introduce the overall pipeline for video scene editing. The pipeline of our \texttt{AudioScenic} framework is depicted in Fig.~\ref{Fig.main}. 
% Following previous work~\cite{Tune, protagonist}, we initialize our model with text-to-image Stable Diffusion and inflate the 2D U-Net structure to process 3D video sequences. Each U-Net is mainly composed of two blocks: Residual block and Transformer block. In the fine-tuning stage, we input the source video and its audio, fine-tuning the specified network parameters through reconstruction loss. During the inference stage, we utilize new audio clips as conditions to edit the video scene content. In particular, we obtain a noisy latent of source video $V$ through the DDIM inversion process $\mathcal{I}(\cdot)$ without auditory condition. Then we employ a new audio clip $A^{*}$ to guide its DDIM sampling process $\mathcal{S}(\cdot)$ through the fine-tuned model. That is,

We introduce the video scene editing pipeline, which is depicted in Fig.~\ref{Fig.main}. 
We employ Stable Diffusion~\cite{latentdiffusion, hierarchical}, which is composed of a VAE autoencoder and a U-Net. First, a VAE Encoder $\mathcal{E}(\cdot)$ compresses the source video into latent feature $\bm{z}_0$, which can be reconstructed back to video by a VAE Decoder $\mathcal{D}(\cdot)$. Second, we modify the U-Net to incorporate audio conditions and train it to remove the noise using the objective:
\begin{equation}\small
    \mathcal{L} = \mathbb{E}_{\mathcal{E}(\bm{x}_0), y, \epsilon\sim\mathcal{N}(0, \mathit{I}), t}\left\lbrack 
\lVert \epsilon - \epsilon_\theta(\bm{z}_t, t, \bm{E}_a, \bm{M}_a, \bm{F}_a) \rVert_2^2 \right\rbrack.
\end{equation}
During inference, we obtain a noisy video latent feature $\bm{z}_t$ through the DDIM inversion process $\mathcal{I}(\cdot)$ without auditory condition. Then we employ a new audio clip $A^{*}$ to guide its DDIM sampling process $\mathcal{S}(\cdot)$ through the fine-tuned model. That is,
% \begin{equation}\small
% \begin{aligned}
% V^{*}=D(\text{DDIM-samp}(\text{DDIM-inv}(\mathcal{E}(V)),A^{*})).
% \end{aligned}
% \end{equation}
\begin{equation}\small
\begin{aligned}
V^{*} &= \mathcal{D}(\mathcal{S}(\mathcal{I}(\mathcal{E}(V)),A^{*})).
\end{aligned}
\end{equation}
% In addition, we employ ControlNet~\cite{controlnet} to preserve the foreground details of the source video during editing.
%In the diffusion model, time $t$ is embedded as time embedding.

% Previous text-guided video editing works often leveraged cross-attention blocks to introduce textual conditions. However, we have found this approach to be unsuitable for audio, as shown in Fig. \ref{Fig.man-failure} (b). The underlying reason may stem from the inability of audio and images to generate semantically aligned attention maps. Consequently, we explore a new way to incorporate audio conditions. The injection of the audio condition into the model is that we add the audio semantic embedding with timestep embedding, which is derived from the sampling timestep $t$. Then we take the fused timestep embedding as input to the U-Net for auditory guidance.

The modified U-Net is mainly composed of stacking Residual blocks and Transformer blocks, as shown in Fig. \ref{Fig.main}. The Residual block incorporates the temporal convolution layers and the Magnitude Modulator module. The Magnitude Modulator module is integrated between temporal convolution layers. In the Transformer block, the Frequency Fuser module is added between the text-video cross-attention layer and the feedforward network. Note that though we are not using the textual condition as input, we retain the text-video cross-attention layer and input blank textual strings to maintain the ability of the pre-trained diffusion model. Temporal convolution layers are utilized to enforce the model to capture temporal information of video latent features. The Magnitude Modulator module achieves temporal visual control through audio magnitude features $\bm{M}_a$. The Frequency Fuser facilitates the video latent features $\bm{z}_t$ in acquiring the frequency information encapsulated within the audio frequency feature $\bm{F}_a$. 

% The modified U-Net is mainly composed of stacking Residual blocks and Transformer blocks. The Residual block incorporates the temporal convolution layers and the Magnitude Modulator module. Temporal convolution layers are utilized to enforce the model to capture temporal information of video latent features. Between convolution layers, we integrate the Magnitude Modulator module, achieving temporal visual control through audio magnitude features. In the Transformer block, we add the Frequency Fuser module between the text-image cross-attention block and feedforward network. The Frequency Fuser facilitates the video latent features in acquiring the frequency information encapsulated within the audio condition. Note that though we are not using the textual condition as input, we retain the text-image cross-attention block and input blank textual strings to maintain the ability of the pre-trained diffusion model.

% \noindent\textbf{Compared to exitsing text-driven video editing methods, \texttt{AudioScenic} has advantages in two aspects.} 1) Text-driven editing methods may face challenges in using manually added text to describe the video scene content, we alleviate this limitation by introducing the audio condition, which naturally matches video in the wild. 2) Since a semantic label includes multiple audio clips in audio datasets and each clip generates a different scene, we can effectively produce rich and diverse video scene content through audio compared to text descriptions.

\subsection{Temporal-Aware Audio Semantic Injection}
\label{sec: audiocondition}
Previous text-driven video editing works~\cite{fatezero, Tune, videop2p, xing2024tokenflow, bai2024uniedit} often leveraged cross-attention layers to introduce conditional guidance. 
These methods rely on the ability of textual descriptions and video to generate semantically matched attention maps through pretraining. Some works~\cite{hierarchical, gen1, protagonist} add CLIP image embedding with timestep embedding to introduce indescribable image conditions. This method enforces the image embedding to apply a greater impact compared to the textual conditions utilized in the cross-attention layers. We observe that the audio conditions and image conditions share numerous similarities. Audio encompasses a range of low-level information that is difficult to convey through textual descriptions. Furthermore, there is a necessity to devise a method whereby audio can exert a powerful impact on videos to generate scenes effectively. Motivated by these considerations, we follow~\cite{hierarchical, protagonist} and propose a temporal-aware audio semantic injection (TASI) process. The injection of the audio semantic into the U-Net is that we add the audio semantic embedding $\bm{E}_a$ with timestep embedding $temb$, which is derived from the sampling timestep $t$. Then we take the timestep embedding fused with audio semantics as input to the U-Net for auditory guidance.

\subsection{SceneMasker for Foreground Texture} 
\label{scenemasker}

In this section, we propose SceneMasker to preserve foreground content along with ControlNet. During scene editing, integrating audio conditions impacts the foreground texture. The core of this issue lies in the generative process, where audio conditions are added with timestep embedding, influencing all regions of the video including the foreground. Therefore, the texture of the foreground is altered in response to audio information. Prior approaches like ControlNet~\cite{controlnet} focus on preserving the spatial structure, specifically the shape of foreground elements, but do not address the nuanced impact of audio conditions on texture consistency. Refer to \S\ref{sec:ablation} for visualization. Consequently, using ControlNet is insufficient to preserve foreground content.
% As shown in Fig. \ref{Fig.man-failure} (c), it is evident that the texture of the man's clothing changes in conjunction with variations in the background. The reason lies in the semantic information of the audio influencing the generation of foreground content.

% ControlNet~\cite{controlnet} has been widely employed to provide spatial control in visual synthesis. However, we have observed that the exclusive use of ControlNet is inadequate for fully preserving the foreground. ControlNet can only maintain the shape of the foreground, without ensuring the consistency of texture. As shown in Fig. \ref{Fig.man-failure} (c), it is evident that the texture of the man's clothing changes in conjunction with variations in the background. The reason lies in the semantic information of the audio influencing the generation of foreground content.

To tackle this specific challenge, we employ SceneMasker, a mask blending module, to constrain the influence of audio conditions only on the scene areas. In concrete, we obtain a binary mask $\mathcal{M}_k$ that splits the foreground and scene content apart. Then the timestep embeddings fused with and without audio embedding are blended with the binary mask. That is,
\begin{equation}\small
\begin{aligned}
temb = \mathcal{M}_k \odot temb_{uf} + (1-\mathcal{M}_k) \odot temb_{f},
\end{aligned}
\end{equation}
where $temb_{f}$ and $temb_{uf}$ denote the timestep embedding fused with and without audio embedding, respectively. $temb$ is the blended timestep embedding used for the proceeded denoising process. SceneMasker complements ControlNet by safeguarding against unwanted texture changes in the foreground, ensuring that both shape and texture remain consistent. This solution offers a more holistic approach to maintaining the integrity of visual content.

\subsection{Magnitude Modulator for Audio-Aligned Temporal Dynamics}
\label{magnitude modulator}
In this section, we introduce the Magnitude Modulator module that is designed to control the visual effects based on audio magnitude. Prompt-to-Prompt~\cite{prompttoprompt} showcases an efficient approach for image editing through the weight control of the cross-attention map between text and images. Building on this foundation, previous work~\cite{aadiff} employs the audio magnitude to modulate the generated video effects by multiplying the smoothed magnitude value with the attention map between target text tokens and the video. However, they rely on an additional textual condition and a specific mapping algorithm to match the text token that mirrors the semantics of the input audio. Worse, this dependency poses an extra challenge in that the required textual token with exact semantics may not always be present in the prompt, thus constraining the adaptability of this approach in video editing scenarios.

To address these limitations, we integrate the Magnitude Modulator to modulate the timestep embedding in the Residual blocks. At each denoising step, we multiply the preprocessed audio magnitude feature $\bm{M}_a$ with the timestep embedding $temb$, expressed as: 
\begin{equation}\small
\begin{aligned}
temb^{'} \!=\! \bm{M}_a \odot f(temb),
\end{aligned}
\end{equation}
$f(\cdot)$ is MLP. We add this weighted timestep embedding $temb^{'}$ with video latent features $\bm{z}_t$ to achieve modulation of the editing effect.

In the inference stage, we can facilitate audio magnitude to control the editing effect of the video. For example, the video scene will change more drastically when the magnitude signal is strong. Our proposed Magnitude Modulator extends the temporal visual diversity of video scenes as we can generate rich scenes within the variance of audio magnitude.

\subsection{Frequency Fuser for Temporal Consistency}
\label{frequency fuser}
We employ audio frequency information to preserve video temporal consistency. The integration of audio frequency information during video editing is crucial. Noteworthy information within audio conditions often resides in the frequency domain~\cite{lin2024echotrack}.
In many real-world scenarios, changes in both visual and auditory elements often coincide. 
Utilizing the frequency characteristics of audio facilitates the visual elements in capturing audio's temporal aspects, thereby producing videos with greater temporal coherence. In addition, one modality may contain noise or irrelevant data absent in the other. Merging audio and video in the frequency domain enables the refinement or filtering of visual frequency components based on audio data. This aspect has not been extensively explored.

We introduce a novel approach through the Frequency Fuser module, which leverages audio frequency characteristics as controlling weights applied to video frequency features within the frequency domain. Initially, we obtain the audio frequency feature $F_a$ in the audio preprocessing stage and transform it into a controlling weight using a Weight Encoder $Enc(\cdot)$. That is, $\bm{F}_{a}^{w} = Enc(\bm{F}_{a} )$.
Note that the $Enc(\cdot)$ includes two MLPs and a non-linear layer.

Drawing inspiration from ~\cite{wu2023freeinit}, we convert video spatial features $\bm{z}_t$ to the frequency domain, preserving the low-frequency components to retain the video's global layout. Subsequently, we integrate the audio frequency weight $\bm{F}_{a}^{w}$ into the high-frequency components to enhance the content details and temporal consistency. The operations are defined as:
\begin{equation}\small
\begin{aligned}
\bm{F}^{L}_{z} &= \mathcal{FFT}_{3D}(\bm{z}_t) \odot \mathcal{P}, \\
\bm{F}^{H}_{z} &= \mathcal{FFT}_{3D}(\bm{z}_t) \odot (1-\mathcal{P}) , \\
\bm{F}^{H'}_{z} &= \bm{F}^{H}_{z} * \bm{F}_{a}^{w}, \\
\bm{z}_t^{'} &= \mathcal{IFFT}_{3D}(\bm{F}^{L}_{z} + \bm{F}^{H'}_{z}),
\end{aligned}
\end{equation}
where $\mathcal{FFT}_{3D}$ denotes the Fast Fourier Transformation operated on video spatial features, $\mathcal{IFFT}_{3D}$ is the Inverse Fast Fourier Transformation that maps back the video frequency features. $\mathcal{P}$ is the spatial-temporal Low Pass Filter. Note that we reshape the $\bm{F}_{a}^{w}$ to align with the video frequency features for effective multiplication.

Through the Frequency Fuser module, we ensure that video features capture the frequency information from the audio, aligning the visual content with the auditory component over time. Such cross-modal frequency components significantly contribute to the temporal coherence of the edited video.

\section{Experiment}
%In this section, we present implementation details and experimental results.

\begin{figure*}[t]
  \centering
  \includegraphics[width=1.0\textwidth]{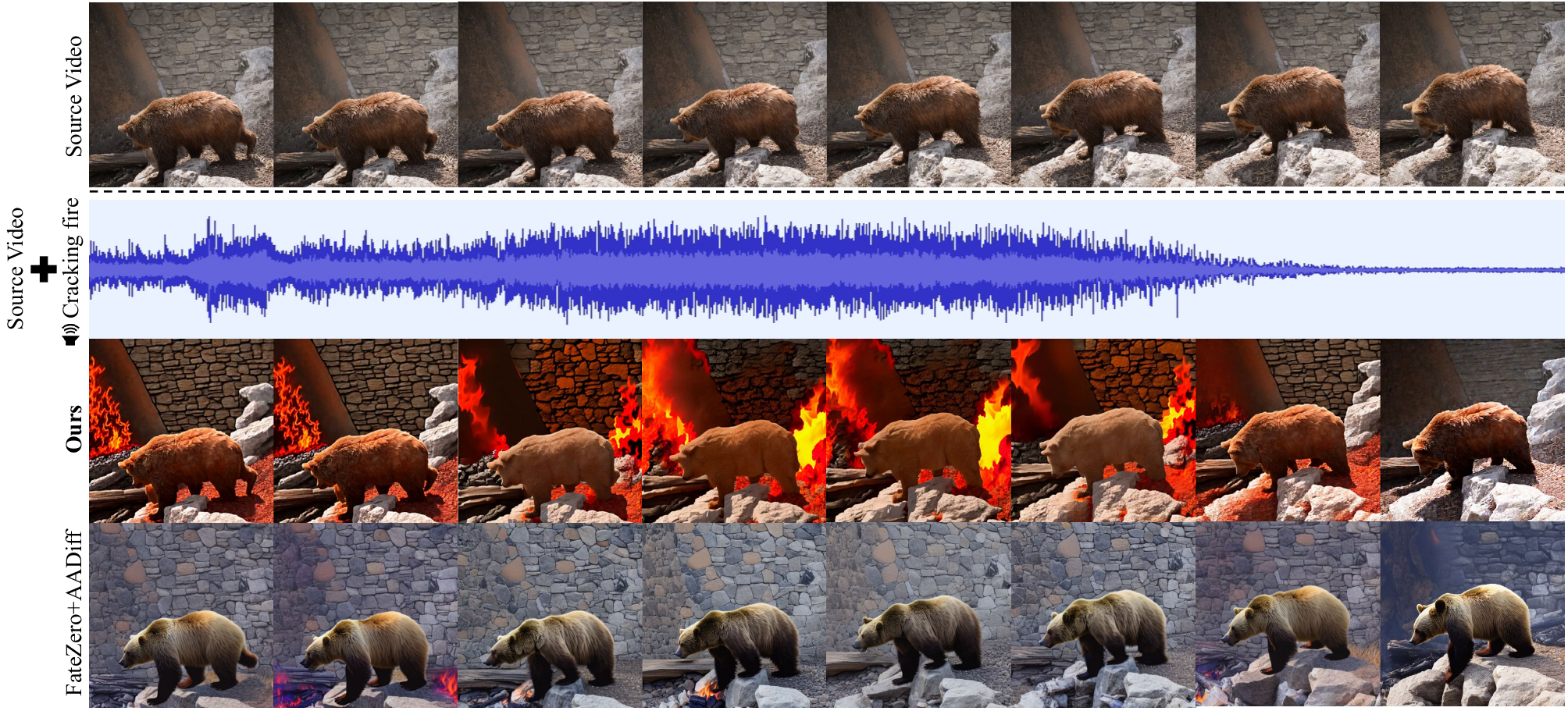}
  \caption{Comparison with baselines through magnitude control (\S\ref{sec: baseline}). The semantic label of input audio is `` Cracking fire''.}
  \label{Fig.magnitude comparison}
  \vspace{-10pt}
\end{figure*}

\begin{figure}[t]
  \centering
  %\fbox{\rule[-.5cm]{0cm}{4cm} \rule[-.5cm]{4cm}{0cm}}
  \includegraphics[width=0.49\textwidth]{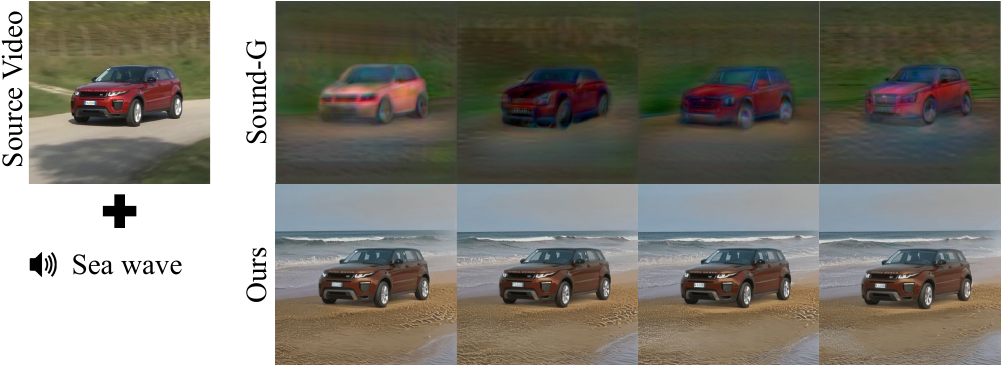}
  % \captionsetup{font=small}
  \caption{Qualitative comparison with Sound-G~\cite{soundguided} (audio-driven image editing method) (\S\ref{sec: baseline}). The semantic label of input audio is ``Sea wave''.}
  \label{Fig.soundguided}
  \vspace{-10pt}
\end{figure}

\subsection{Implementation Details}
Our \texttt{AudioScenic} is implemented based on Stable Diffusion. We initialize the diffusion model and inflate the 2D U-Net structure to process 3D video input. In the fine-tuning stage, we fine-tune the model on a source video at a resolution of $768 \times 768$ and then conduct inference to edit videos with different scenes. We fine-tune the model for 300 steps with a learning rate of $3 \times 10^{-5}$ and employ the DDIM sampler with 50 steps. We benefit from ImageBind and Mel Spectrogram for the feature extraction of audio. During inference, we employ ControlNet to preserve the shape consistency of foreground content. We use Grounded-SAM and XMem to obtain the binary mask for the SceneMasker module. Detailed descriptions of these models can be found in the Appendix. We experimented on a single NVIDIA RTX 3090 GPU. Our code will be released.

% \noindent\textbf{Datasets.} Following previous work~\cite{xing2024tokenflow, Tune, fatezero, soundini, protagonist, videop2p, imagebind}, we apply our method to selected videos from DAVIS dataset and Audioset dataset to conduct qualitative and quantitative experiments. Audio clips are sampled from Audioset and ESC-50 datasets. Refer to the Appendix for detailed descriptions of datasets. 

\subsection{Datasets}
\textbf{DAVIS~\cite{davis2017}.} The DAVIS dataset is designed for the task of video object segmentation. In this dataset, the main objects are segmented in the scene and divided by semantics. We select video samples from DAVIS dataset for training and inference.

\noindent\textbf{Audioset~\cite{audioset}.} The Audioset dataset contains videos from YouTube annotated into 527 classes. For each class, the dataset contains an unbalanced training set, a balanced training set, and an evaluation set. We select videos and audio clips from the Audioset balanced training sets and evaluation sets for training and inference.

\noindent\textbf{ESC-50~\cite{esc50}.} The Environment Sound Classification (ESC) dataset contains 2000 5s audio clips. It has 50 audio classes, including object and natural sound. We choose audio clips from this dataset for generating video scenescapes in the inference stage.

\subsection{Evaluation Metrics} 
\label{sec: metrics}
In this section, we introduce three metrics for quantitative measure. Following~\cite{clip, fatezero, wang2004ssim}, we use Semantic Accuracy(Sem-A) to measure the editing semantic accuracy and Structural Similarity Index Measure (SSIM) to validate the foreground content degradation. In addition, we propose a new Temporal Score (Temp-S) to measure the temporal consistency comprehensively.

\noindent\textbf{Sem-A~\cite{fatezero}.} Sem-A is the percentage of frames where the edited image has a higher CLIP-T score (average similarities between the CLIP embedding of conditions and all frames) to the target condition than the source condition. It measures the performance of editing methods at the semantic level. Note that we use the textual semantic labels of audio clips to compute the CLIP-T Score and compare the result with the text-driven and audio-driven methods.

% \noindent\textbf{ImageBind Score~\cite{imagebind}.} To evaluate the semantic alignment of results, we use the semantic labels of audio as scene editing targets and measure the similarity score between the video and semantic labels in the ImageBind feature space.
\noindent\textbf{SSIM~\cite{wang2004ssim}.} SSIM measures the similarity between two images. We focus on editing video scenes while keeping foreground content unaltered. Therefore, we employ a mask to isolate the foreground regions of the video and subsequently calculate SSIM for these regions before and after editing. Through this metric, we quantify video foreground quality degradation.

\noindent\textbf{Temp-S.} When measuring the temporal consistency of edited videos, using the CLIP-F score (average pairwise similarities of the CLIP embedding of images) alone does not provide a comprehensive evaluation, as the CLIP-F score can yield high scores even when the video is not edited at all. Consequently, we introduce Temporal Score (Temp-S) for a holistic assessment, that is, Temp-S = CLIP-F * CLIP-T, where CLIP-T calculates average similarities between the CLIP embedding of conditions and all frames. Temp-S evaluates the temporal consistency of results on the premise of semantically accurate scene editing.

% \begin{figure}[t]
%   \centering
%   %\fbox{\rule[-.5cm]{0cm}{4cm} \rule[-.5cm]{4cm}{0cm}}
%   \includegraphics[width=0.48\textwidth]{picture/multi-semantic-class-crackingfire.pdf}
%   % \captionsetup{font=small}
%   \caption{\textbf{Qualitative results.} Audio clips corresponding to the same semantic label ``Cracking fire'' generate diverse scenes.}
%   \label{Fig.multi-semantic}
%   \vspace{-10pt}
% \end{figure}

\begin{figure}[t]
  \centering
  %\fbox{\rule[-.5cm]{0cm}{4cm} \rule[-.5cm]{4cm}{0cm}}
  \includegraphics[width=0.48\textwidth]{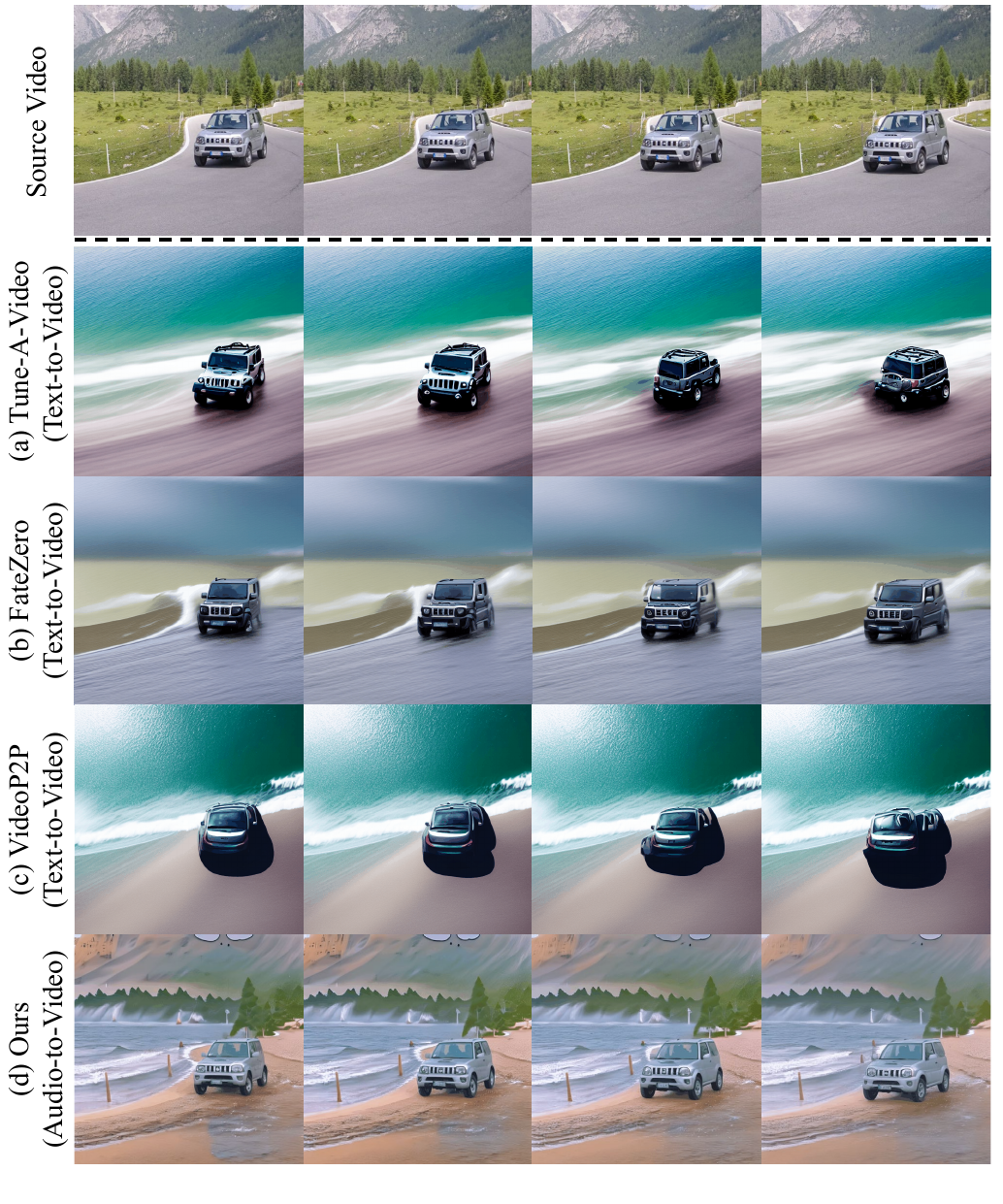}
  % \captionsetup{font=small}
  \caption{Qualitative comparison (\S\ref{sec: baseline}). We compare results between (a-c) text-driven methods~\cite{Tune, fatezero, videop2p} and (d) ours. For (a-c), the input text is ``a jeep car is driving beside the sea wave''. For ours, the semantic label of audio is ``Sea wave''.}
  \label{Fig.baseline-half}
  \vspace{-10pt}
\end{figure}

\begin{table}[t]
    % \vspace{-10pt}
	\fontsize{6}{8}\selectfont
	\resizebox{0.48\textwidth}{!}{
        \setlength\tabcolsep{4pt}
		\renewcommand\arraystretch{1.1}
    	%\begin{center}
    		\begin{tabular}{m{1.9cm}<{\raggedright} || m{1.0cm}<{\centering}  | m{1.1cm}<{\centering} | m{1.1cm}<{\centering}}
    		\hline\thickhline
            \rowcolor{mygray}
             Model & Sem-A$\uparrow$ & SSIM$\uparrow$ & Temp-S$\uparrow$ \\ 
    			\hline\hline
            Tune-A-Video~\cite{Tune} & 0.3875 & 0.8611 & 27.31 \\
            FateZero~\cite{fatezero} & 0.4375 & 0.8944 & 28.48 \\
    	    VideoP2P~\cite{videop2p} & 0.525 & 0.8263  & 28.38 \\
           \hline
            \textbf{Ours} & \textbf{0.825} & \textbf{0.9403}& \textbf{29.31} \\
			\hline
    		\end{tabular}
	}
    % \captionsetup{font=small}
    \caption{Quantitative results of ours and text-driven video editing methods (\S\ref{sec: baseline}). The results are measured on ten samples selected from DAVIS~\cite{davis2017} and Audioset~\cite{audioset}. Sem-A, SSIM, and Temp-S denote Semantic Accuracy, Structural Similarity Index Measure, and Temporal Score respectively (\S\ref{sec: metrics}).}
	\label{tab:quantity}
 \vspace{-15pt}
\end{table}

\begin{figure}[t]
  \centering
  %\fbox{\rule[-.5cm]{0cm}{4cm} \rule[-.5cm]{4cm}{0cm}}
  \includegraphics[width=0.48\textwidth]{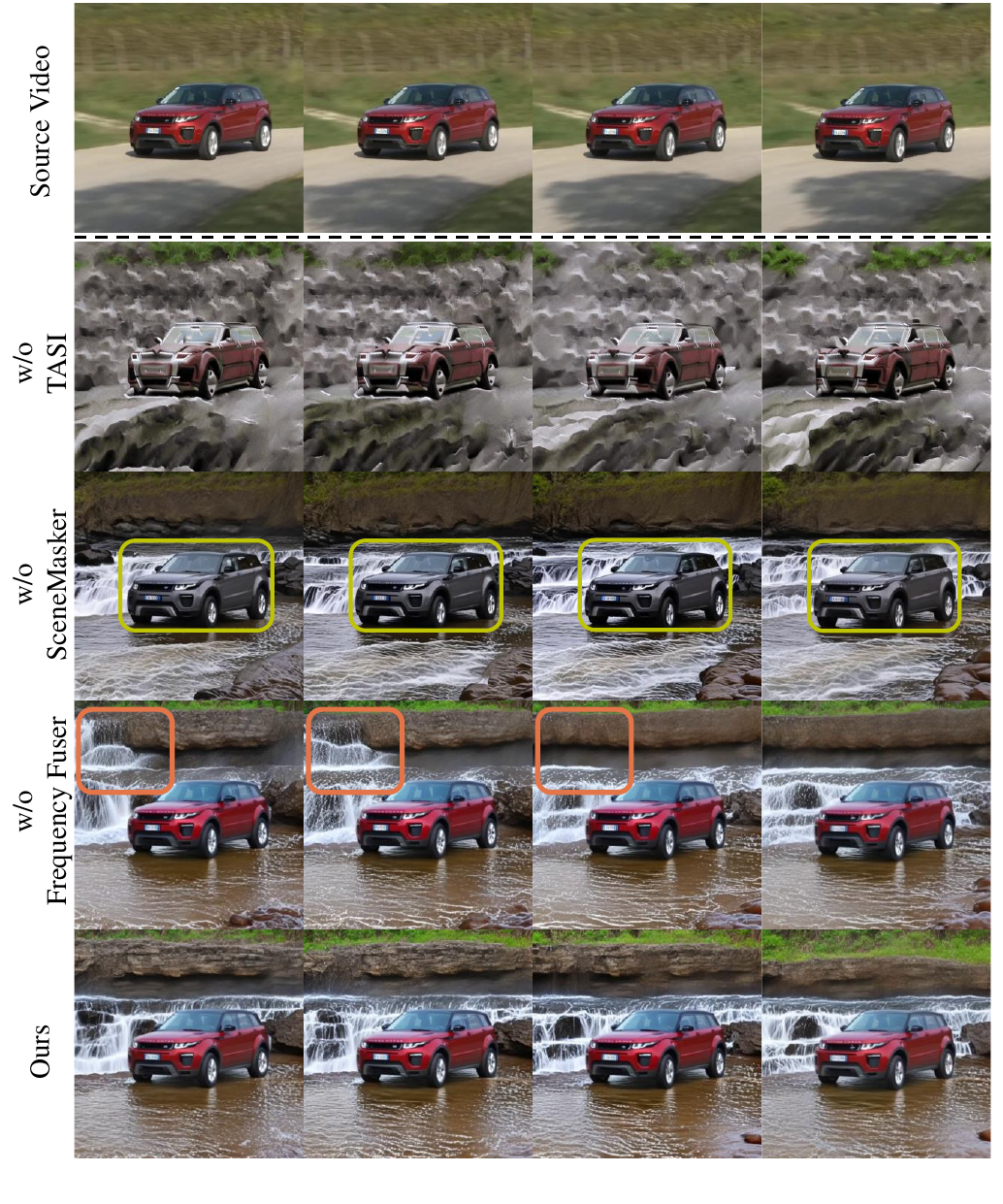}
  % \captionsetup{font=small}
  \caption{Qualitative ablation studies (\S\ref{sec:ablation}). The semantic label of input audio is ``Waterfall''. \textcolor{yellow}{Yellow} boxes indicate the textual change of the car. \textcolor{orange}{Orange} boxes indicate the temporal inconsistency of the scene. Zoom in for details.}
  \label{Fig.Ablation}
  \vspace{-10pt}
\end{figure}

\subsection{Comparisons}
\label{sec: baseline}
\textbf{Competing Methods.} We show our superiority in foreground preservation, temporal dynamics, and temporal consistency through qualitative and quantitative evaluation. To ensure a fair comparison, we select two sets of competing methods: \textbf{i) Text-driven methods}: 1) Tune-A-Video~\cite{Tune} fine-tunes an inflated diffusion model on a single video to produce similar content. 2) FateZero~\cite{fatezero} utilizes Prompt-to-Prompt in the video editing task and uses textual descriptions to change the value of the text-image cross-attention map, facilitating video editing. 3) VideoP2P~\cite{videop2p} designs a Text-to-set model to generate a set of semantically consistent videos and uses a decoupled-guidance strategy to improve the model's robustness. \textbf{ii) Audio-driven methods}: Sound-G~\cite{soundguided} trains an audio encoder to produce audio features aligned with image features. It then manipulates images with audio features using StyleGAN~\cite{karras2020stylegan}.  

% \textit{ii) Audio-driven Baseline:} Soundini~\cite{soundini} employs a user-provided mask to achieve local region editing with audio conditions. Soundini is limited to achieving local region editing, which is different from scene editing. When it comes to global scene editing, Soundini relies on an extra image cloning technique to fuse the foreground and the edited scene. We compare our results with Soundini adopting the image cloning technique. Note that as Soundini has not released the source code, we conduct qualitative and quantitative comparisons using the ``Underwater bubbling'' sample presented on its project page.  

%\subsubsection{Qualitative Comparisons}

\noindent\textbf{Qualitative Comparisons.} We first compare the results of our model and text-driven methods in semantic scene editing. The qualitative results are depicted in Fig.~\ref{Fig.baseline-half}. The text-driven methods are given the prompt ``A jeep car is moving beside the sea wave'' and our audio-driven framework takes a ``Sea wave'' semantic audio clip as input. The edited results from Tune-A-Video flickers and VideoP2P are weak in producing a vivid sea wave scene. FateZero preserves temporal consistency but it is weak in expressing temporal dynamics of the scene. The scenes generated by FateZero remain static. All the competing methods fail to maintain the shape and texture of the car. Our method produces a more coherent and dynamic scene of sea waves. Besides, our results retain the shape and texture of the car.

We compare our results with the Audio-driven methods, Sound-G, in semantic scene editing. As shown in Fig. \ref{Fig.soundguided}, the videos generated by Sound-guided fail to present ``Sea wave'' scenes, with a noticeable loss of detail in the car, and a lack of temporal dynamics across frames. In contrast, our approach not only preserves the texture of the car but also introduces dynamic and coherent scenes.

We compare with previous work in temporal dynamic scene editing through magnitude control. We establish a baseline that combines AADiff~\cite{aadiff} and FateZero~\cite{fatezero}, controlling the editing effect by employing audio magnitude to control the value of text-video cross-attention maps. We qualitatively compare our results with this baseline in Fig.~\ref{Fig.magnitude comparison}. We employ an audio clip of ``Cracking fire'' to guide the scene editing. It can be observed that while the baseline method is capable of generating visual effects of flames in certain frames, it fails to modulate these visual effects in accordance with the variations in the audio magnitude. The more noticeable visual dynamics aligned with audio magnitude variance shows the priority of our method.

\begin{table}[t]
    % \vspace{-10pt}
	\fontsize{6}{8}\selectfont
	\resizebox{0.48\textwidth}{!}{
        \setlength\tabcolsep{4pt}
		\renewcommand\arraystretch{1.1}
    	%\begin{center}
    		\begin{tabular}{m{1.9cm}<{\raggedright} || m{1.0cm}<{\centering}  | m{1.1cm}<{\centering} | m{1.1cm}<{\centering}}
    		\hline\thickhline
            \rowcolor{mygray}
             Model & Sem-A$\uparrow$ & SSIM$\uparrow$ & Temp-S$\uparrow$ \\ 
    			\hline\hline
            w/o TASI & 0.25 & 0.9047 &25.11 \\
            w/o SceneMasker & 1.0 & 0.8863 &29.41 \\
    	  w/o Frequency Fuser & 1.0 & 0.9467  & 28.18 \\
           \hline
            \textbf{Ours} & 1.0 & \textbf{0.9512}& \textbf{29.61} \\
			\hline
    		\end{tabular}
	}
    % \captionsetup{font=small}
    \caption{Quantitative results of ablation study (\S\ref{sec:ablation}). ``w/o TASI'' means without using temporal-aware audio semantic injection to integrate audio semantics and using audio-video cross-attention layers instead.}
	\label{tab:ablation}
 \vspace{-15pt}
\end{table}

% \begin{table}[t]
% 	\fontsize{6}{8}\selectfont
% 	\resizebox{0.47\textwidth}{!}{
%         \setlength\tabcolsep{4pt}
% 		\renewcommand\arraystretch{1.1}
%     	%\begin{center}
%     		\begin{tabular}{m{1cm}<{\centering} || m{1.7cm}<{\centering} | m{1.5cm}<{\centering} | m{1.5cm}<{\centering} | m{1.2cm}<{\centering} }
%     		\hline\thickhline
%             \rowcolor{mygray}
%              Metric  & Tune-A-Video~\cite{Tune}  & FateZero~\cite{fatezero} & VideoP2P~\cite{videop2p} & \textbf{Ours}\\
%             \hline
%     		 Realness$\downarrow$ & 3.0 & 2.6 & 3.3 & \textbf{1.1} \\
% 			\hline
%               Coherence$\downarrow$ & 3.5 & 1.9 & 3.2 & \textbf{1.4} \\
% 			\hline
%     		\end{tabular}
% 	}
%     \caption{\textbf{Quantitative results} of our User Study.}
% 	\label{tab:userstudy}
%  \vspace{-12pt}
% \end{table}

\begin{figure}[t]
  \centering
  %\fbox{\rule[-.5cm]{0cm}{4cm} \rule[-.5cm]{4cm}{0cm}}
  \includegraphics[width=0.48\textwidth]{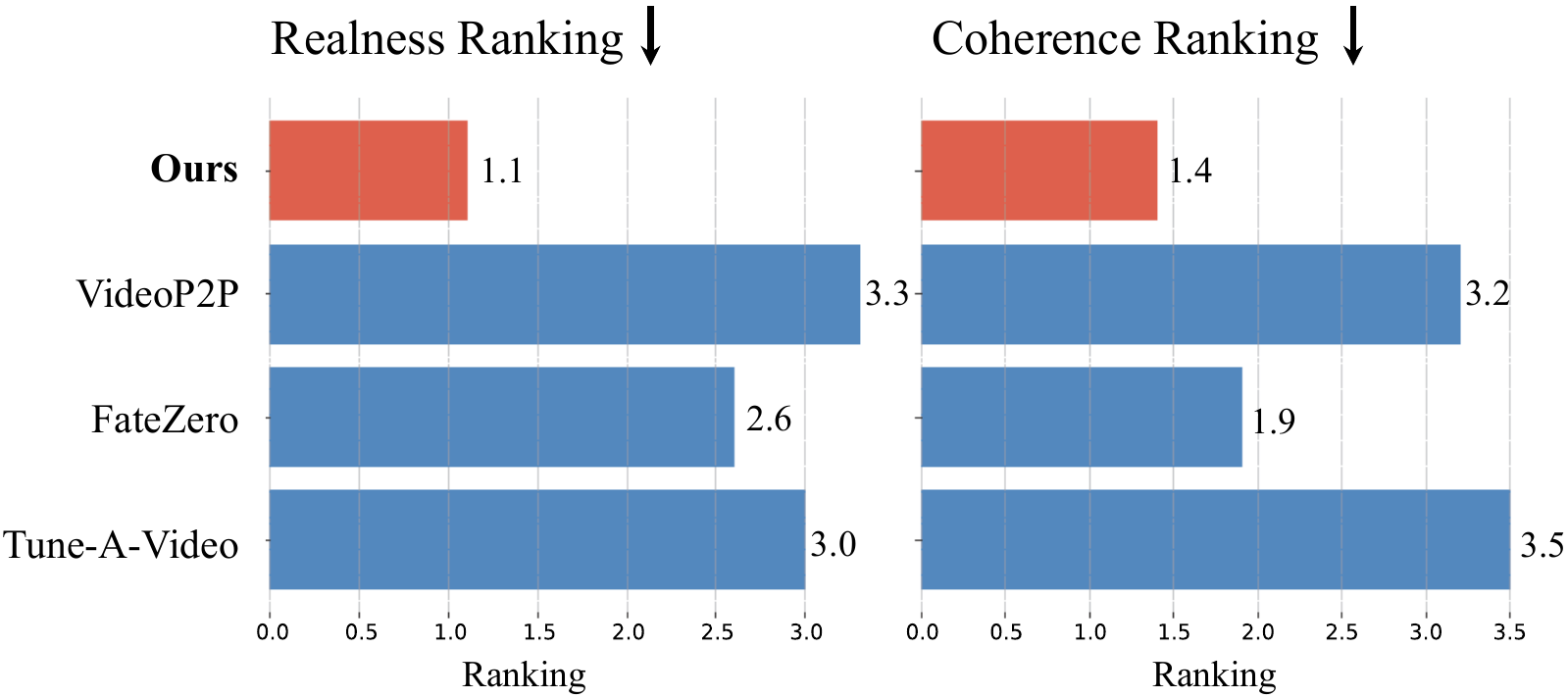}
  % \captionsetup{font=small}
  \caption{Quantitative user studies (\S\ref{sec: baseline}). Ranking 1 is best and Ranking 4 is worst. The results are measured on ten samples selected from DAVIS~\cite{davis2017} and Audioset~\cite{audioset} datasets. A lower ranking indicates greater preference.}
  \label{Fig.userstudy}
  \vspace{-10pt}
\end{figure}

\noindent\textbf{Quantitative Analysis.} We first compare our method with the text-driven methods using ten sets of comparisons. The results are shown in Tab.~\ref{tab:quantity}. The Sem-A score of our model is 0.825, which is significantly higher than the next-best score by VideoP2P at 0.525. This vast gap of nearly \textbf{36.3\%} suggests that our model has a deeper understanding of the conditions, leading to editing results that better reflect the semantically matched scenes. In addition, our model stands out with an SSIM score of 0.9403, lifting the score \textbf{8.42}\%, \textbf{4.88}\%, and \textbf{12.12}\%, compared to Tune-A-Video, FateZero, and VideoP2P respectively. This significant lead demonstrates our model's remarkable ability to keep the texture and shape of foreground content during editing. It shows that our model can edit scenes without losing the video's foreground appeal. The Temp-S score of our model promotes \textbf{6.82}\%, \textbf{3.15}\%, and \textbf{3.17}\%, over Tune-A-Video, FateZero, and VideoP2P respectively. This superiority illustrates our model's capacity to ensure temporally smooth transitions and consistency over time. This is pivotal for maintaining realness and preventing visual flicking in edited videos.

\noindent \textbf{User Study.} To provide a complete measure of the quality of our method and text-driven methods, we conduct a set of human evaluation experiments, based on 10 pair-wise comparisons. Concretely, 10 academics are asked to respectively rank (1 is best and 4 is worst) different results based on two aspects (1) Realness ~\textit{Please select the more realistic results}, and (2) Coherence ~\textit{Which results are more temporally consistent}. The results are shown in Fig. \ref{Fig.userstudy}. the \textbf{average rankings} of our method are \textbf{1.3} and \textbf{1.4} in content realness and temporal coherence, respectively, receiving the highest preference ratings from users and showing the superiority of \texttt{AudioScenic}. 

% \noindent\textbf{Quantitative Analysis.} We first compare our method with the text-driven baselines using ten sets of comparisons. The results are shown in Tab.~\ref{tab:quantity}. According to the \textbf{Sem-S} metric, our method achieves the best Sem-A (36.3\% better than the second-best method, VideoP2P), which means our results are more visually correlated with target semantics. In addition, our results achieve the highest \textbf{SSIM} and \textbf{Temp-S} (2.8 \% and 14.7\% better than the second-best method, FateZero, respectively). These demonstrate that our method generates more temporally consistent results and keeps the foreground unchanged.  As for \textbf{User Study}, the average rankings of our method are 1.3 and 1.4 in content realness and temporal coherence, respectively, receiving the highest preference ratings from users. Tab. \ref{tab:soundini} shows the quantitative results between ours and Soundini. We express superiority in the aspects of semantic accuracy and temporal consistency.

\subsection{Ablation Study}
\label{sec:ablation}
To verify the necessity of each component in our method, we qualitatively conduct ablation studies on the \textbf{i)} temporal-aware audio semantic injection (TASI), \textbf{ii)} SceneMasker module, and \textbf{iii)} Frequency Fuser module. The semantic label of input audio is ``Waterfall''. The qualitative result is in Fig. ~\ref{Fig.Ablation} and the corresponding quantitative result is in Tab. \ref{tab:ablation}. 

\noindent\textbf{The Effect of TASI.} To test the function of TASI, we eschew its use and instead employ audio-video cross-attention layers to integrate audio semantics for video editing guidance. The quantitative results show a drop in \textbf{75}\% in Sem-A score. The visual changes in Fig. \ref{Fig.Ablation} also showcase the semantical mismatch. The model fails to present waterfall scenes. These results show that our TASI is effective and significant in introducing audio semantics into the model.

\noindent\textbf{The Effect of SceneMasker.} Through qualitative visualization, without the SceneMasker module, the color of the car changes (from red to black) before and after editing. In the quantitative assessment, the \textbf{6.82}\% drop of SSIM score also showcases the change of the car. These results emphasize the crucial role of the SceneMasker module in foreground preservation during editing.  

\noindent\textbf{The Effect of Frequency Fuser.} As shown in Fig. \ref{Fig.Ablation}, the waterfall scene shows visual flicking between the first and third frames. In Tab. \ref{tab:ablation}, we observe the \textbf{4.83}\% drop of Temp-S score for editing results without Frequency Fuser. These results indicate that deactivating the Frequency Fuser module degrades the temporal consistency of edited scenes and video quality.

% To verify the necessity of each component in our method, we qualitatively conduct ablation studies on the temporal-aware audio semantic injection process, SceneMasker module, and Frequency Fuser Module. The qualitative result is in Fig. ~\ref{Fig.Ablation} and the semantic label of input audio is ``Waterfall''. First, using audio-video cross-attention blocks to inject audio conditions into the model rather than adding audio embedding with timestep embedding fails to generate waterfall scenes. Second, without the SceneMasker module, the car color turns black since the audio embedding impacts car regions. Third, deactivating the Frequency Fuser module degrades the temporal consistency of waterfall scenes and video quality. We also conduct quantitative ablation studies in Tab. \ref{tab:ablation}. The method that uses audio-video cross-attention to inject audio conditions shows low performance in all metrics. The models without SceneMasker and without Frequency Fuser both achieve perfect Semantic Accuracy but vary in their SSIM and Temp-S scores, indicating that removing these components affects aspects of foreground preservation and temporal consistency, respectively. With the integration of all components, our method showcases remarkable performance in qualitative and quantitative assessments.

\begin{figure}[t]
  \centering
  %\fbox{\rule[-.5cm]{0cm}{4cm} \rule[-.5cm]{4cm}{0cm}}
  \includegraphics[width=0.49\textwidth]{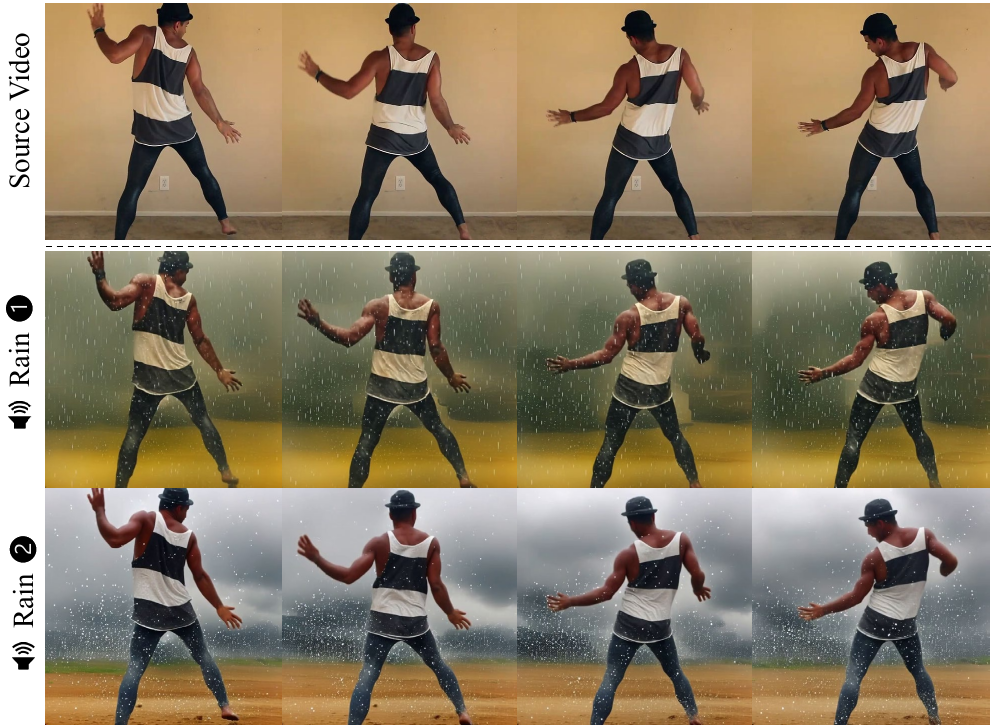}
  % \captionsetup{font=small}
  \caption{ Audio clips corresponding to the same semantic label ``Rain'' generate diverse scenes.}
  \label{Fig.multu-class}
  \vspace{-10pt}
\end{figure}

\subsection{Applications of \texttt{AudioScenic}}
\textbf{Audio-Guided Semantic Editing.} We show the editing results of our method in Fig.~\ref{Fig.figure1}. Given source videos (first row), \texttt{AudioScenic} supports diverse and coherent scene editing with the corresponding audio variation (second to fourth rows). We also achieve scene-style transitions using emotional audio clips, like happy music and sad music. We can employ audio clips belonging to the same semantic label to generate different visual scenes, as shown in Fig. \ref{Fig.multu-class}.
Refer to the Appendix for more visualization.

\noindent \textbf{Magnitude-Controlled Temporal Scene Dynamics.} Magnitude is one of the crucial properties of audio. \texttt{AudioScenic} controls the temporal scene dynamics based on the change in audio magnitude. As shown in Fig.~\ref{Fig.magnitude comparison}, our method can produce visual changes in the scene aligned with the audio magnitude, improving the temporal visual dynamics. Given the audio clips with ``Cracking fire'' semantics, the change of scene content becomes noticeable as the value of the magnitude variances along the time axis. 
%Besides, in Fig.~\ref{Fig.multi-semantic}, we show that multiple audio clips of the same semantic label can generate different video scenes.

% To verify the necessity of each component in our method, we qualitatively conduct ablation studies on the SceneMasker module, Frequency Fuser Module, and ControlNet in the denoising sampling. The semantic label of input audio is ``Waterfall''. The results are depicted in Fig.~\ref{Fig.Ablation}. First, our model removing ControlNet fails to preserve the car shape and makes its motion inconsistent. Second, without the SceneMasker module, the car color turns white since the audio embedding causes a global influence on the frames. Third, deactivating the Frequency Fuser module degrades the temporal consistency of scenes and video quality. With the integration of all components, our method showcases remarkable performance in qualitative assessments.

\section{Conclusion}
In this paper, we introduce \texttt{AudioScenic}, an audio-driven framework designed for video scene editing. 
\texttt{AudioScenic} integrates audio semantics into the visual scene through a temporal-aware audio semantic injection process. We further introduce a SceneMasker module that maintains the integrity of the foreground content.
\texttt{AudioScenic} exploits the audio magnitude and frequency to control the temporal dynamics and enhance the temporal consistency. 
We present a new metric named temporal score for more comprehensive validation of temporal consistency. Finally, We demonstrate the effectiveness of our method. 
%through extensive experiments, showing significant improvements over existing text-driven and audio-driven models in video scene editing.

% In this work, we introduce \texttt{AudioScenic}, an audio-driven framework for video scene editing. \texttt{AudioScenic} leverages the unique properties of audio to guide the editing process. Our framework employs audio embeddings as a condition to edit the video scenes. We present Magnitude Modulator, an audio magnitude-aware module for controlling the editing effect, enabling a wide range of editing diversity. Additionally, we integrate a Frequency Fuser module to improve the temporal coherence of edited scenes. Our approach improves the visual diversity of the scenes and preserves temporal consistency. Finally, We demonstrate the effectiveness of our method through extensive experiments, showing significant improvements over existing text-driven and audio-driven models in video scene editing.

\bibliographystyle{ACM-Reference-Format}
\bibliography{main}

\end{document}